# Automated screening of privacy risks in medical AI

Konstantinos Vrettos[1], Eleftherios Tzanis[1,2], Haridimos Kondylakis[3,4], Evangelia E. Vassalou[1,5], Florence X. Doo[6,7], Michail E. Klontzas[1,2,4,5,#]

1. Artificial Intelligence and Translational Imaging (ATI) Lab, Department of Radiology, School of Medicine, University of Crete, Heraklion, Crete, Greece
2. Division of Radiology, Department of Clinical Science Intervention and Technology (CLINTEC), Karolinska Institute, Huddinge, Sweden
3. Computer Science Department, University of Crete, Crete, Greece
4. Institute of Computer Science, FORTH, Greece
5. Department of Medical Imaging, University Hospital of Heraklion, Crete, Greece
6. University of Maryland Institute for Health Computing, North Bethesda, MD, USA
7. Department of Diagnostic Radiology and Nuclear Medicine University of Maryland, Baltimore, MD, USA

#Corresponding author

**Address for correspondence**

Michail E. Klontzas, MD, PhD

Assistant Professor of Radiology

Artificial Intelligence and Translational Imaging (ATI) Lab

Department of Radiology, School of Medicine,

University of Crete

Voutes, 71003, Heraklion, Crete, Greece

Tel: +30 2811391351 E-mail: miklontzas@gmail.com; miklontzas@ics.forth.gr; miklontzas@uoc.gr

ORCID: 0000-0003-2731-933X

## Abstract

**Medical artificial intelligence (AI) models are often shared without their training data, assuming this protects the patients who contributed them. However, a model can reveal whether a patient's records were used in training, a privacy breach known as a membership inference attack. This risk is usually reported as a cohort average, which can appear negligible while individual patients are almost certainly exposed. Identifying potentially vulnerable records before developing a production model could make privacy assessment more practical. We measured each patient's vulnerability to membership inference attack, across 7,789 models trained on six clinical datasets in four modalities, including two adapted pretrained medical models. Training choices changed how many patients had high attack scores, but the patients who ranked highest by risk largely stayed the same across the configurations we tested. Stronger weight decay reduced the exposed cohort up to threefold, yet largely the same patients remained most at risk across training recipes, architectures and initializations, including foundation models. Therefore, we developed a lightweight screening tool that uses four small models to identify patients with high risk of exposure before production-model training. We also assessed if flagged patients could be protected without removing their data. Privacy screening should become incorporated into data preparation for clinical AI.**

## Main

Medical AI models are routinely shared and deployed without their training data, but withholding records alone does not necessarily ensure patient privacy[1]. Membership inference attacks illustrate this limitation. By querying a trained model, an attacker can infer whether a particular patient's record was part of its training set[2–4]. In clinical settings, membership alone can reveal sensitive information, such as a diagnosis, when the record can be linked to a person and inclusion in the training cohort is inherently informative[5].

Membership-inference risk is usually summarized across the training cohort, and near-chance average attack performance has been read as evidence of privacy[6]. Evaluations at low false-positive rates[3] and among the most vulnerable records[7] show that such averages conceal substantial individual risk variations. A patient-level audit of clinical models identified patients whose membership can be inferred almost with certainty, with very high attack discrimination and documented demographic disparities[5]. However, whether an inexpensive screen can identify the same high-scoring patients consistently and across various clinical training configurations remains a practical open question. Theory links memorization to atypical examples[8,9], whereas empirical studies show that memorization changes during training[10], with the composition of the training set[3] and with regularization[2,4,11].

The distinction has practical consequences. If vulnerability is largely determined by patient data, patients at risk could be identified before a production model is trained. If vulnerability instead changes with each training configuration, risk would need to be assessed after each model is built and the resulting models would require their own privacy audits. Early predictors of memorization exist[12,13], but they read the training trajectory of the model being evaluated and were developed on non-clinical benchmarks. They also do not test whether the same vulnerable records remain among those most at risk across different training approaches. We therefore asked whether a data screening method could identify these patients before production-model training.

Here we evaluate membership-related signals in a study comprising 7,789 trained models across six clinical datasets spanning four modalities: chest radiography, dermoscopy, electrocardiography and magnetic resonance imaging (MRI). The models range from small networks trained from scratch to fine-tuned foundation models. We found that average attack performance conceals substantial variation between patients  and that training choices change how many patients exceed a specified risk threshold. Patient rankings by risk persisted across recipes, architectures, initializations and foundation model adaptations, and their records' held-out predictive difficulty (atypicality) helped anticipate the most at-risk patients. This makes risk predictable before training. We translated these findings into an open-source lightweight screening tool, which uses four small proxy models before production-model training to flag at-risk patients and assessed whether the flagged patients could be made less vulnerable to membership inference during training without excluding their records.

## Cohort average risks conceal patient-level exposure variation

To understand what drives membership-inference vulnerability, a measurement for each patient is first needed. We used six public datasets: CheXpert[14] and NIH-CXR14[15] (chest radiographs), ISIC-2020[16] (dermoscopy), PTB-XL[17] (12-lead electrocardiograms), SCOPE-MRI[18] (shoulder MRI) and PneumoniaMNIST[19] (paediatric chest radiographs) (Fig. 1b). Cohorts comprised 546 to 18,619 patients with an average of 1.2 to 16.0 records each, where each record represents one model input: a radiograph, dermoscopic image, MRI series, or an electrocardiogram. For each training configuration (arm), multiple models were trained per dataset (Methods) each on a random half of the patients and we verified that every patient was included in the training set of half of the models. Such models trained to probe membership are known as shadow models[3]. For each record, the attack AUC is the probability that a model trained on that patient is more confident in the record's true label than a model that was not trained on that patient (0.5, chance; 1.0, complete separation). An AUC of 0.5 indicates chance ordering, whereas 1.0 indicates complete separation of member and non-member confidence values in the evaluated models. A patient's risk score was the highest attack AUC among their records, scores above 0.7 were considered exposed. This threshold was prespecified as an operational cutoff, drawing on the conventional lower bound of acceptable discrimination for clinical prediction models[20], but is not a validated boundary for privacy harm (Methods). PneumoniaMNIST lacks patient identifiers and was analysed per example (1 example = 1 record).

Across the five datasets with patient identifiers, 49–95% of patients exceeded the risk-score threshold (in PneumoniaMNIST, which lacks patient identifiers, 21% of examples were exposed). These proportions substantially exceeded those expected by chance (Fig. 1c, grey lines). By contrast, the population attack AUC calculated by pooling all records ranged from 0.56 to 0.93. Even on ISIC-2020, where this overall AUC was only 0.56, approximately half of the patients exceeded the patient-level threshold, illustrating variation concealed by the cohort average. These statistics summarize different comparisons and are not probabilities of patient disclosure. As a positive control for detecting memorization, we randomized the labels of a set of records per dataset (canaries, Methods), which a model can fit only by memorizing, so a working attack must score them as members. Canaries met the detection criterion in every reference arm (median AUC 0.86–1.00, against 0.54–0.94 for ordinary records; Fig. 1d). Arms that did not meet this criterion were excluded from the comparisons, while canary records were excluded from all reported scores. Failure to detect canaries is inconclusive because it can reflect less memorization or a mismatch between the control and the training procedure.

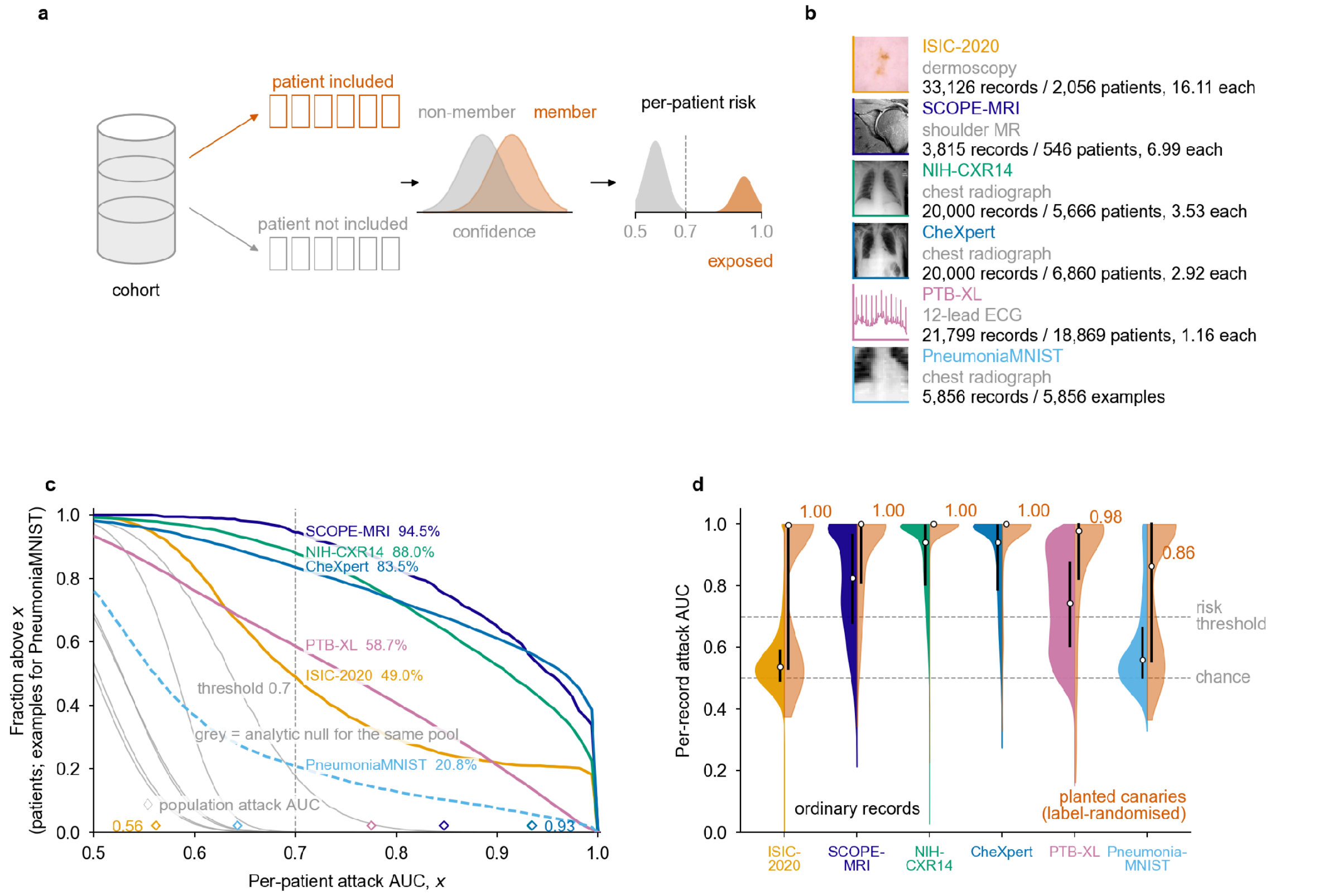

**Fig. 1 Patient-level auditing reveals widespread variation concealed by cohort averages. a**, Audit design. For each training configuration (arm), 32–256 shadow models were each trained on a random half of the patients. Every patient was therefore in the training set of half of the models (members) and absent from the rest (non-members). For each record, the confidence of member and non-member models was compared to calculate the attack AUC. A patient's risk score is the maximum over their records, and patients with risk >0.7 are classified as exposed. **b**, Datasets, modalities, and sample sizes, including canaries. **c**, Fraction of patients whose risk score exceeds each per-patient attack AUC under each dataset's reference recipe (PneumoniaMNIST, examples; dashed). Grey curves represent fractions expected by chance alone if training on a patient made no difference to a model's confidence, given the numbers of member and non-member models and each patient's number of records. Percentages indicate the observed fraction of patients with per-patient attack AUC >0.7. Open diamonds, population-level attack AUC (0.56–0.93). **d**, Positive control. Per-record attack AUC of ordinary records (left half of each violin) and planted records with randomized labels. White dots, medians; black bars, interquartile ranges; numbers, median canary AUC. Comparisons are made at matched model counts (Methods).

## Training changes how many patients exceed the risk threshold

We next investigated how training choices changed the measured vulnerability. To compare their effects with run-to-run variation, we compared each reference training strategy with a training that changed one setting and with a seed replicate of the reference strategy. Increasing weight decay, a standard regularization that penalizes large model weights, from $10^{-4}$ to $10^{-2}$ reduced the at-risk fraction 1.38- to 2.95-fold on the four datasets where both arms passed the canary control (PneumoniaMNIST, CheXpert, ISIC-2020, NIH-CXR14 ). Reseeding changed the number by only 1.07-fold (0.18–3.02 percentage points; Fig. 2a,b). On NIH-CXR14, at-risk patients decreased from 4,767 to 1,615 (out of a total of 5,416 patients). Of 19,750 records, 17,701 had significantly lowered attack AUCs, whereas only 36 had significantly increased (Fig. 2d). On SCOPE-MRI the count was near saturation, with 95% of patients above the threshold, so we compared the fraction above it after subtracting what chance alone would give and the results showed that weight decay lowered this from 0.69 to 0.51, against 0.011 between seed replicates. Weight decay therefore lowered the measured risk on all five datasets where the comparison could be made, though by varying amounts (Fig. 2c). Not all interventions had the same effect though, as on PneumoniaMNIST, mixup increased the number above threshold by 38% and shrinking the training set to a quarter of the reference size, increased the percentage of patients at risk by 33% (Fig. 2e).

Improving performance in the under-fitted ISIC-2020 reference model was accompanied by a similar reduction in measured vulnerability. The ISIC-2020 reference model, trained on 224-pixel thumbnails, performed worse than published alternatives, so we retrained it following a competition-winning training approach[21]. Held-out AUROC increased by 0.174 (95% confidence interval (CI) 0.159–0.190), and the population attack's true-positive rate at a 0.1% false-positive rate fell 13-fold, which is the low-false-positive-rate operating point recommended for privacy audits[3]. The fraction above the risk threshold fell from 70.8% to 45.2% of 2,056 patients (difference, 25.6 percentage points; 95% CI 23.2–28.1; $N$ = 32 models), against 1.2 percentage points between seed replicates (Fig. 3a,c). Across the 512-pixel arms, higher vulnerability was positively correlated with the gap between training and held-out performance (Spearman $\rho = 0.79$, $p<0.05$; Fig. 3d). The better-performing model therefore left fewer patients at risk, but still 45% of them scored above 0.7 (attack AUC >0.7).

We extended the analysis to pretrained medical foundation models. ECGFounder[22] (Net1D, 30.7 million parameters, pretrained on about 10 million electrocardiograms) was fine-tuned on PTB-XL. The fraction above the risk-score threshold rose with how much of the model was fine-tuned. When only the final classification layer was trained (5,125 parameters, 0.017% of the total), 22.5% of patients exceeded the threshold. When the last few layers were also trained (3.42 million parameters, 11%), this fraction was 30.4% and when the whole model was fine-tuned, it reached 54.6% (Fig. 3e). The fraction also increased with training duration (Fig. 3b), as under full fine-tuning, it increased from 10.9% after two epochs to 54.6% after twenty. Stronger weight decay reduced it by 2.7–4.5 percentage points, close to ten times the shift between seed replicates. On SCOPE-MRI, MRI-CORE[23], a foundation model pretrained on MRI and adapted through low-rank adaptation (LoRA; Methods), achieved held-out AUROC 0.767, compared with 0.708 for the ImageNet-initialized ResNet-18 and 0.746 for a vision transformer, and a slightly lower fraction above threshold (89.6% against 95%). The linear-probe comparison was inconclusive because it did not meet the canary criterion (canary AUC 0.58). Fine-tuning all weights lowered held-out AUROC to 0.728 and raised the fraction above threshold to 96.3%, while setting weight decay to $10^{-2}$ yielded AUROC 0.724 and a fraction of 95.2% .

Most of these reductions in the at-risk fraction required negligible additional GPU energy, with some even lowering it. We measured GPU-board energy with the built-in counter which excluded CPU and storage. Weight decay and data augmentation changed energy per epoch by 0.2–5.5% and held-out AUROC by −0.021 to +0.038 (Fig. 2f). Freezing most of ECGFounder and training only its final layers, cut training energy by 59–61% and the fraction above the risk-score threshold from 55% to 22–30%(Fig. 3e). On SCOPE-MRI, adapting MRI-CORE with LoRA used 16% less training energy than fine-tuning all its weights (10.1 against 12.0 kWh per 32 models) with fewer patients above threshold (89.6% against 96.3%) and higher AUROC (0.767 against 0.728). At 512 pixels, a ResNet-18 used 65% less energy than an EfficientNet-B3 and had fewer patients above threshold (45.2% against 55.1%; difference of 9.9 percentage points; 95% CI 7.7–12.1). The 512-pixel, competition-winning configuration had a lower fraction above the risk threshold than the 224-pixel configuration (45.2% against 70.8%), but required 8.5 times the energy. Conversely, mixed-precision training cut energy by 71–72% on PTB-XL and NIH-CXR14 with similar diagnostic performance. Thus, training defined how many patients were at high attack risk, but it remained unclear whether it was the same patients who remained most at risk, even when training changed.

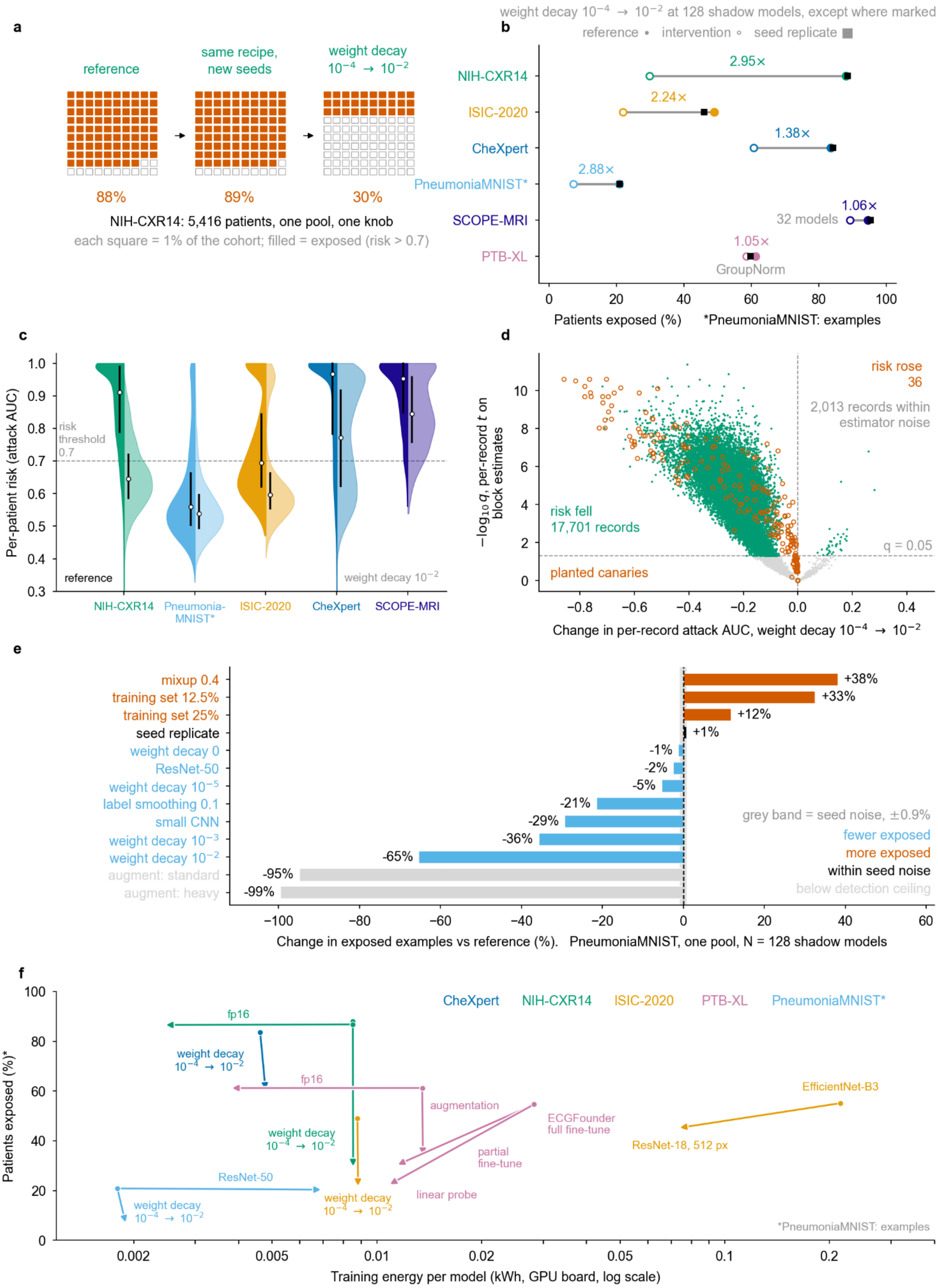

**Fig. 2 Training changes how many patients exceed the risk threshold. a**, A single cohort (NIH-CXR14, 5,416 patients). Each square represents 1% of patients, and filled squares mark scores >0.7. Retraining the reference recipe with new random seeds had negligible impact on the fraction above threshold (from 88% to 89%), whereas increasing weight decay from $10^{-4}$ to $10^{-2}$ reduced it to 30%. **b**, Fraction above threshold fraction under the reference recipe (filled circle), a seed replicate (black square) and the modified recipe

(open circle), with fold changes. On PTB-XL, whose weight-decay arm did not meet the canary criterion, the contrast is batch normalization replaced by group normalization. **c**, Per-patient risk scores under the reference training strategy (left, solid) and under weight decay (right, pale). White dots, medians; black bars, interquartile ranges. **d**, Per-record change in attack AUC on NIH-CXR14 under weight decay. Two-sided t-tests on eight independent block estimates per condition, with Benjamini–Hochberg correction across 19,750 records. Vermillion circles, canaries; ordinary-record result counts exclude canaries. **e**, All recipes evaluated on PneumoniaMNIST, ranked by the change in the number of examples above threshold . Blue, fewer ; vermillion, more; black, within the ±0.9% seed band; grey, data augmentation arms that modified the training data and were outside the scope of this work. **f**, GPU-board energy versus the fraction of patients above the 0.7 threshold. Each arrow runs from a reference arm (dot) to a modified arm that changes one setting/hyperparameter (arrowhead), in the dataset's colour; x, measured training energy per model (log scale); y, fraction above the risk threshold at the final checkpoint. Vertical arrows indicate changes in the at-risk percentage with little change in measured GPU-board energy (for example, at weight decay $10^{-4} \rightarrow 10^{-2}$), horizontal arrows indicate changes in energy with little change in the percentage of at-risk patients (mixed precision, ResNet-50) and diagonal arrows change both (freezing most of ECGFounder, EfficientNet-B3 → ResNet-18 at 512 pixels on ISIC-2020). GPU-board energy from the NVML counter on one RTX 5000 (PUE 1.0); SCOPE-MRI is omitted because it was measured on a different GPU. $N$ = 128 shadow models, except SCOPE-MRI and the PTB-XL row in b ($N$ = 32, due to computational and energy considerations).

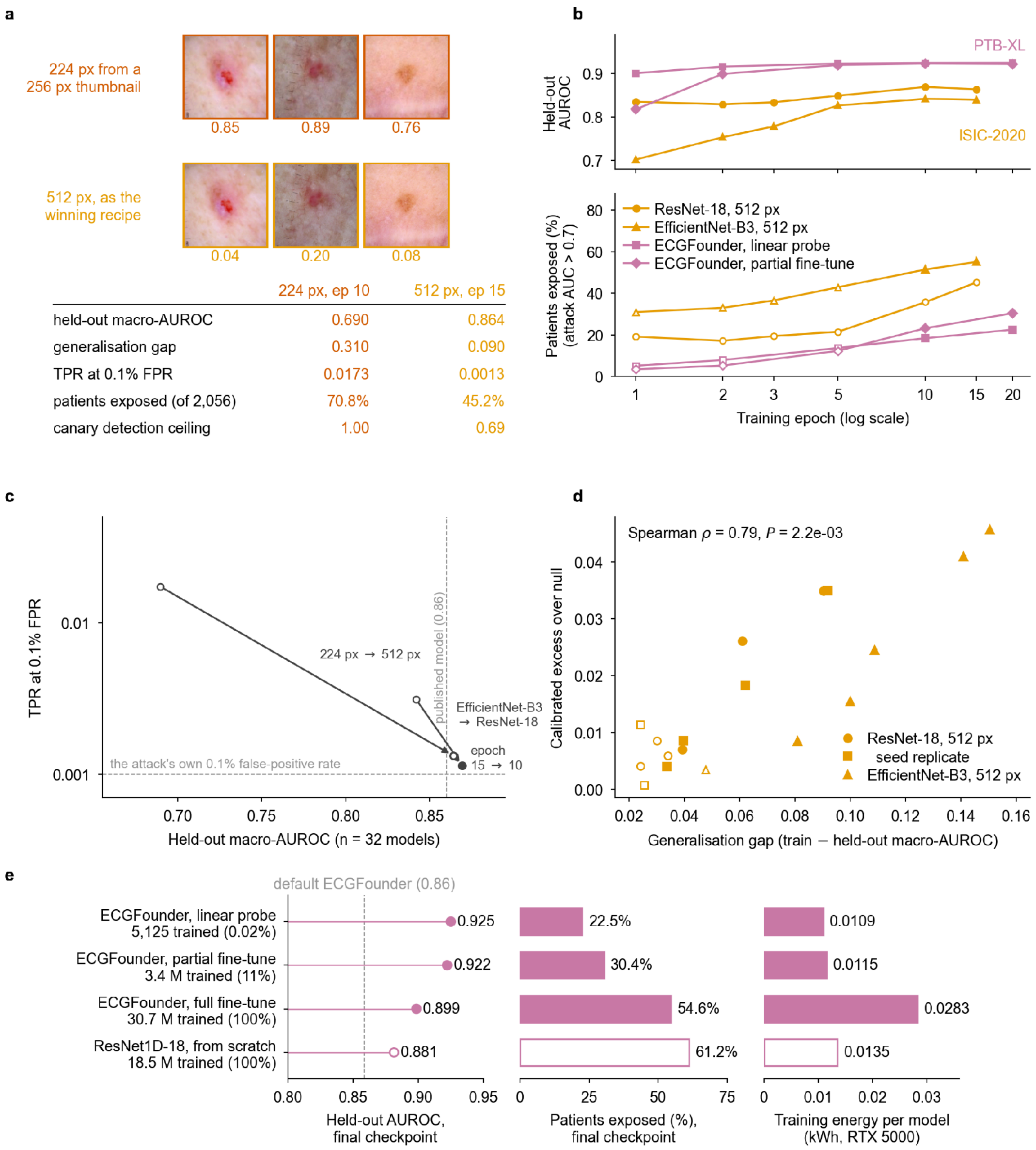

**Fig. 3 Diagnostic performance and attack risk under different training configurations .** Panels a, c and d use a single ISIC-2020 pool (2,056 patients; $N = 32$ shadow models), in which the training approach (resolution, epochs, regularization), architecture and checkpoint were varied. **a**, The three records whose attack AUC decreased most under the competition-winning approach: 224 pixels (upper) and 512 pixels from the full-resolution archive (lower). Numbers indicate the attack AUC of each record. **b**, Held-out AUROC (upper) and fraction above the risk threshold (lower) by epoch for the two 512-pixel ISIC-2020 models and for ECGFounder trained as a linear probe or partially fine-tuned on PTB-XL ($N = 32$). Open markers, checkpoints with median canary AUC below 0.7. **c**, Changes in diagnostic and attack performance (open markers, before; filled markers, after). Dashed lines, published model AUROC (0.86) and the population attack's 0.1% false-positive rate, shown for reference. **d**, Generalization gap against calibrated excess over null for the three 512-pixel variations (ResNet-18, seed replicate of ResNet-18 and EfficientNet-B3) at six checkpoints. Spearman $\rho = 0.79$, $p<0.05$; across the $n = 12$ checkpoint-arm

combinations that met the canary criterion (filled markers). **e**, y: ECGFounder on PTB-XL trained as a linear probe or partially or fully fine-tuned (final checkpoint, $N = 32$), x: held-out AUROC, fraction above threshold and training energy per model. Open symbols, ResNet1D-18 trained from scratch on the same pool; dashed line, default ECGFounder (0.86).

## The data determine who is exposed

To distinguish between changes in the number and ranking of exposed patients (above the 0.7 threshold), we compared how different training strategies ranked patients by risk score, relative to the agreement between seed replicates of the same training approach. This normalized stability, S, equals 1 when the rank correlation between recipes matches that between seed replicates and $S \geq 0.8$ was set as the limit for agreement. Under a change of weight decay, S was 0.75–0.98 across the five datasets whose weight-decay arm passed the canary control (all except PTB-XL) (Fig. 4f). On SCOPE-MRI, we compared four alternatives to the reference model, a ResNet-18 initialized from ImageNet. Patient rankings were preserved under a stronger weight decay of $10^{-2}$ (S = 0.92), with a vision transformer instead of a ResNet (S = 0.97) and with additional pretraining on knee MRI[24] (S = 1.00) (Fig. 4f,g). A ResNet-18 trained from random initialization, which shared no pretrained weights with the reference, yielded S = 0.75 (95% CI 0.69–0.80), though its upper bound still reached 0.8. On PTB-XL, the fully fine-tuned ECGFounder ranked patients similarly to the network trained from scratch (S = 0.85, 95% CI 0.84–0.86). The partially fine-tuned and final-layer-only versions placed about half as many patients above threshold but ranked them similarly (correlation with the fully fine-tuned version, corrected for run-to-run noise, 0.78–0.81). This was replicated on SCOPE-MRI, where various fine-tuned versions of another foundation model (MRI-CORE), ranked patients similarly to the ImageNet-initialized ResNet-18 (S = 0.96, 95% CI 0.93–1.00; Fig. 4g). ResNet-18 models trained for 50 instead of 10 epochs yielded S = 0.93–0.98 (ISIC-2020 0.93, 95% CI 0.90–0.96; NIH-CXR14 0.94, 0.94–0.95; CheXpert 0.98, 0.98–0.99; Fig. 4g). Training therefore changed how many

patients exceeded the threshold but largely preserved which patients ranked highest, suggesting that vulnerability is determined by the records themselves.

### Atypicality identifies exposed patients

Atypicality was strongly associated with patient risk scores. It reflects how poorly a record's dataset-provided label is predicted by models that never trained on that patient (the held-out component of a memorization score[8]), and is therefore defined relative to the task, labels, training population and model. Atypicality separated the 5% with the highest risk scores from the rest on all six datasets (Cohen's $d$ = 1.78, 95% CI 1.70–1.86, to 4.18, 4.00–4.40; all $q < 10^{-12}$; Fig. 4c,d). Computing atypicality from an independently seeded replicate arm changed $d$ by at most 7%, reducing concern about shared estimation noise. Alternative predictors were weaker: records per patient yielded $d$ = 0.10–0.94, and disagreement between held-out models yielded $d$ = −1.00 to 2.17. This association was also observed in the higher-performing configurations. On ISIC-2020 at 512 pixels (held-out AUROC 0.864), $d$ was 2.74 (95% CI 2.54–2.96) for a ResNet-18 and 2.82 (2.62–3.05) for an EfficientNet-B3. On SCOPE-MRI (held-out AUROC 0.735–0.746 vs published models at 0.681–0.725; ref. 18), $d$ was 2.92 (2.61–3.34). For ECGFounder (foundation model), $d$ was 2.65, against 2.63 for the network trained from scratch. Atypicality was therefore strongly associated with the patients with the highest risk across every dataset and model type tested, from small networks trained from scratch to fine-tuned foundation models. The question that remained was whether those high scores were distributed evenly across patient subgroups.

Specific patient groups were more likely to exceed the risk threshold . Patients aged 80 years or older were over-represented on all four datasets analysed (CheXpert, NIH-CXR14, ISIC-2020 and PTB-XL), with odds ratios of 1.83–6.45 across all (23) training variations, including fine-tuned ECGFounder (2.44, 95% CI 2.21–2.69). They were also over-represented among the 5% patients with the highest scores in most training variations. Other differences were specific to one dataset. Men were over-represented on ISIC-2020 and PTB-XL (odds ratios 1.32 and 1.31), while patients younger than 40 years were over-represented in the SCOPE-MRI (top-5% tail; risk ratio 3.20, 2.58–3.86). No racial or ethnic group showed statistically significant over-representation in the evaluated contrasts (Black vs White patients on CheXpert, odds ratio 1.01, 95% CI 0.73–1.41). The association with Medicare insurance on CheXpert attenuated after age adjustment (age-adjusted odds ratio 1.12, 0.92–1.36). Therefore, risk was not evenly distributed and patients aged 80 years or older were consistently more likely to exceed the risk threshold, whereas other demographic differences varied between datasets.

### Screening patients before production-model training

. The association between atypicality and high risk scores across training approaches suggested that patients at risk could be identified during data preparation, before any production model is trained. A data holder needs to know which patients are at risk before training, without the computational and energy cost of a full audit. We therefore built DataConfessor, an open-source screening tool which estimates atypicality (Fig. 4a). It trains four small, inexpensive proxy models for one epoch, two on each half of the patients (dataset split in half at the patient-level), and scores every patient with the models that did not see them. It then flags the 20% with the highest atypicality. If two independent pairs of proxies disagree on the ranking (Spearman correlation below 0.7), it withholds predictions and trains for longer. This self-check assesses agreement between the two proxy rankings, and not their accuracy in identifying patients with high audit scores. The proxies and final, audited models were trained separately on the same underlying cohort. We ran the tool once per dataset and evaluated its predictions against the audit of every training configuration on that dataset that passed the canary control criterion.

Across all five datasets with patient identifiers, DataConfessor recovered 72.7-100% of the patients placed in the top 5% by the reference audit, 3.6–5 times the rate of random flagging, in 1.4–7.4 GPU-minutes (Fig. 4e). On SCOPE-MRI, where its self-check required five-epoch proxies, it recovered 96% (95% CI 87–100) and its target was the epoch-5 audit because final-checkpoint scores were saturated. The flags transferred to recipes with different weight decay, normalization or architecture, recovering 65–100% of the top 5%, against 73–100% for reseeded reference recipes. Transfer was weaker on PTB-XL, both for group normalization (recall was 65.1%, against 72.7% for the reference recipe) and for fine-tuned ECGFounder (56–60%, against 65–68% for the network trained from scratch at matched N=32 models). Early epoch top-5% scores overlapped the final top 5% by only 4–10%, whereas atypicality at those epochs recovered 79%. The screen used 0.0015–0.085 kWh of GPU-board energy per dataset, 37–1,200 times less than the full audits used here, depending on model used and number of shadow models (Fig. 4h). The screen thus identified most patients in the audit's highest-risk tail using minimal energy and compute. To place this performance in context, we compared the tool with simpler statistical predictors, calculated directly from the data.

We compared 27–29 statistical predictors per dataset across four datasets (Fig. 5a), including label and metadata rarity, a disclosure-risk score used by national statistical offices[25], the special-uniques disclosure score, image outlier scores and unsupervised image representations. Label rarity was the strongest statistical predictor on three of the four datasets. It came close to the tool on NIH-CXR14 (88.6% against 91.5%) and on ISIC-2020 (99.0% against 100%), where the highest-risk patients were almost all patients with melanoma. However, it recovered only 50.0% of the CheXpert tail and 39.6% of the PTB-XL tail, against 75.2% and 72.7% for the tool (95% CIs: label rarity 44.2–55.2% and 36.5–42.7%; tool 70.3–79.7% and 69.8–75.5%). The best statistical predictor therefore varied between datasets (40–99%), whereas the tool's reference-audit recall was at least 73% and had the highest point estimate in each of these four comparisons. Its value, therefore, lies in that consistency, as a data holder cannot know in advance which cheap statistic will be informative on their own data. Finding the patients with high attack risk also raises the question of how to mitigate their vulnerability.

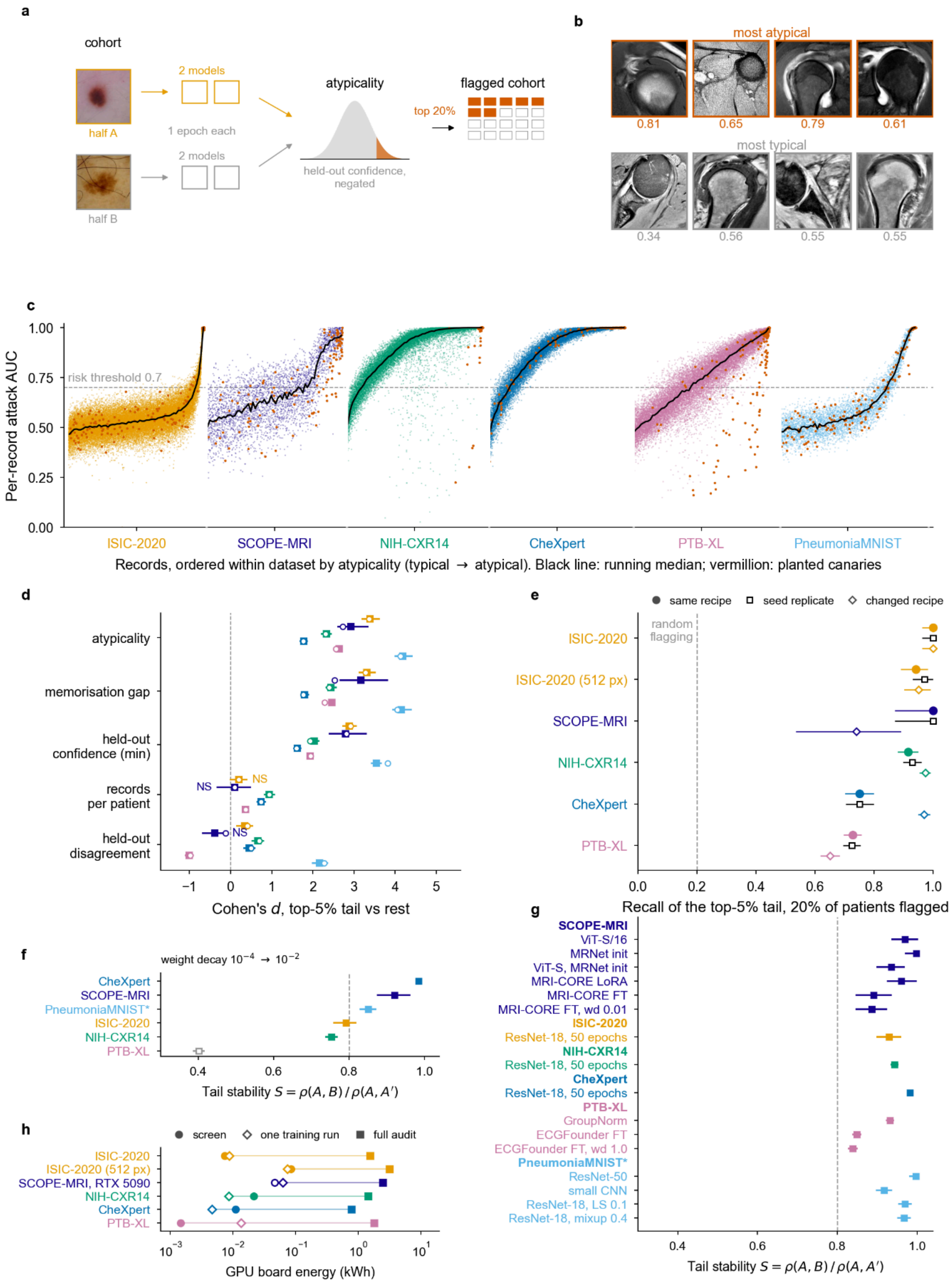

**Fig. 4 Atypicality, a property of the data, supports screening before production-model training. a**, The screening tool. Four proxy models are trained on complementary patient halves, and each scores only the half of patients it was not trained on. One epoch was the default; SCOPE-MRI used five. **b**, The four SCOPE-MRI series with the highest and four with the lowest atypicality within a single class. Numbers,

attack AUC of each series across 32 models. **c**, Records, ordered within each dataset by atypicality and plotted against attack AUC. Black line, running median; vermillion, canaries. **d**, Cohen's *d* for separation of the 5% of patients with the highest attack risk from the remainder, for five candidate predictors. Filled squares, predictors computed from the audited reference models; open circles, predictors computed from an independently seeded replicate arm. NS, not significant after Benjamini–Hochberg correction. **e**, Performance of the screening tool. Recall of the top 5% when the tool flags 20% of patients. For each dataset, one set of flags is scored against the reference training strategy (filled circle), its seed replicate (black square) and modified training strategies (open diamonds: weight decay $10^{-2}$ for ISIC-2020, NIH-CXR14 and CheXpert; group normalization for PTB-XL; EfficientNet-B3 for ISIC-2020 at 512 pixels; weight decay $10^{-2}$ for SCOPE-MRI). Grey dashed line, random-flagging recall (20%). f, Normalized stability *S* of the patient risk ranking under weight decay $10^{-4} \rightarrow 10^{-2}$, relative to each dataset's reference recipe. Dashed line, the 0.8 convention threshold; open grey marker, an arm that did not meet the canary criterion. **g**, *S* under changes of architecture, initialization, normalization, regularization or epoch (LS, label smoothing). **h**, GPU-board energy of the screen (circles), one training run (diamonds) and the full audit (squares), per dataset. Error bars, 95% patient-bootstrap intervals (2,000 resamples).

## Testing mitigation while retaining patient records

Identifying patients with high risk scores is most useful if it can guide protection. Removing their data can shift vulnerability to other records[26] and could also remove the cases a clinical model must learn to recognize. On the 512-pixel ISIC-2020 model, for example, the flagged patients were almost exactly the patients with melanoma (407 of 411, holding 561 of 579 melanoma images). We therefore kept every patient in the training data and changed only how records were learned. The first mitigation strategy was self-distillation, in which the model is trained on predictions from teacher models that never trained on the patient that a record belongs to, rather than on the actual dataset labels[27] (Methods). This approach reduced the fraction of the flagged patients who were above the risk threshold, without sacrificing performance (Figure 5). Across ISIC-2020, PTB-XL and NIH-CXR14, self-distillation lowered flagged patients' risk scores by 0.14–0.34 and unflagged patients by up to 0.20. The other strategies were less effective. Fixed softened labels did not lower the risk scores (mean change, +0.003) and class-wise histogram matching lowered them by −0.20 but reduced AUROC to 0.51 (Fig. 5b,d). No viable mitigation approach reduced the population attack's true-positive rate at a 0.1% false-positive rate (Fig. 5c) (except for histogram matching which brought AUC down to 0.51). Thus, although self-distillation reduced mean audit scores, it did not establish protection against membership inference at low false-positive rates and is an optional feature of DataConfessor.

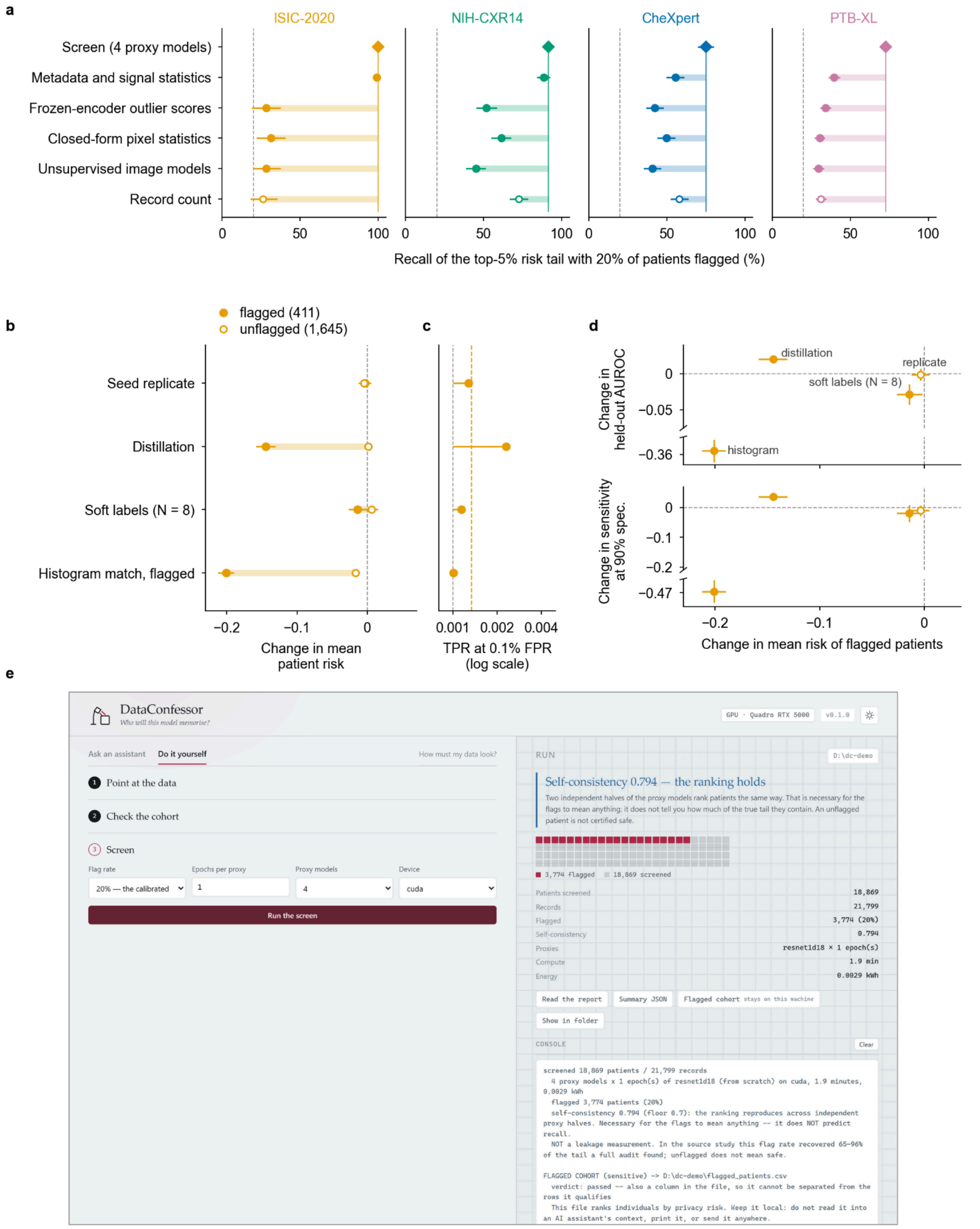

**Fig. 5 Screening compared with statistical predictors, and the trade-off between privacy mitigation and diagnostic performance. a**, Percentage of each dataset's top-5% at-risk tail recovered when 20% of patients are flagged. Diamond and vertical rule, the screen (four proxy models, one epoch each); circles, the best predictor in each statistical predictor family, chosen per dataset from 27–29 predictors (Methods); open circle, record count. Shaded bar, the screen's margin over each predictor; dashed line, random flagging (20%). Error bars, 95% CIs from 2,000 bootstrap resamples of the tail patients. **b**, Change in mean per-patient directional score relative to the undefended 512-pixel ISIC-2020 ResNet-18 (mean

flagged-patient score 0.85), for the 411 flagged patients (filled) and the 1,645 unflagged patients (open). **c**, True-positive rate of the population attack at a 0.1% false-positive rate over all models of each arm (log scale; stems start at chance, 0.001; dashed line, undefended arm, 0.0013). **d**, Change in flagged-patient mean score against change in held-out AUROC (top) and melanoma sensitivity at 90% specificity (bottom), mean over the models of each arm with 95% CIs. Defended arms are compared by paired two-sided *t*-tests; the seed replicate (open) uses Welch's test.**e**, The DataConfessor interface. Error bars, 95% CIs from 2,000 patient bootstrap resamples.Panels b–d use the final ISIC-2020 checkpoint (epoch 15). Per-patient scores are the maximum directional attack AUC over non-canary records.

## Discussion

Membership-inference risk in medical AI has largely been measured on individual trained models and summarized across their cohorts, leaving open whether patient-level vulnerability is specific to a model or to the data. By auditing more than 7,700 models across six datasets and four modalities, from small networks to fine-tuned foundation models, we found that training choices changed how many patients had high audit scores, while largely the same patients remained most at risk across training recipes, architectures and initializations. Since the identity of vulnerable patients appears to be determined by the data, it can be estimated before any production model exists, and privacy assessment can become part of data preparation without relying on extensive, computationally expensive privacy audits of fully trained models.

Our results provide new insights for understanding and predicting patient-level privacy risk. Earlier studies showed that cohort averages understate membership risk[3,7] and that some patients in clinical models can be identified with near-perfect attack discrimination[5]. First, we found that high patient-level attack scores were not confined to a small minority. Between 49% and 95% of the patients in every cohort with patient identifiers exceeded the risk threshold even where the population attack AUC (0.56) was close to chance, which is commonly read as evidence of privacy[6]. Second, although memorization is known to change during training, with dataset composition and with regularization[2–4,10,11], changes in regularization, architecture, initialization, and model family altered mainly how many patients had high scores, while largely preserving the identity of patients who ranked highest . This is also what distinguishes atypicality from early-training predictors of memorization[12,13], as those approaches require information from the model being trained, whereas atypicality can be estimated before the production model exists. Here, we show that a lightweight proxy screen can anticipate much of this patient ranking across variable training choices, including adaptations of pretrained medical foundation models.

Learning theory offers a plausible explanation for this. As in the case of long-tailed data, records that are unconventional or poorly represented by the rest of a dataset may require more memorization in order to fit them[8,9], and atypicality captures how poorly a record can be predicted from other patients. Training then influences how much of that memorization occurs. Attack risk increased with the generalization gap, the number of fine-tuned parameters and decreased with stronger weight decay. In light of this, the data help rank patients by how difficult their records are to predict from the rest of the cohort (atypicality), while training determines how strongly those differences are expressed in the final model. This is consistent with the number of patients above the audit threshold changing while their ranking remained relatively stable, with better-fitted models leaking less and with the observation that removing the most vulnerable records can expose the next layer[26]. It also helps explain why simpler statistics computed directly from the data performed inconsistently: atypicality is specific and defined relative to the clinical task, so a record can look unusual yet be easy to predict, or look ordinary yet be hard to predict. Because the data largely determine who is at risk, that risk is not shared equally and in the cohorts studied, patients aged 80 years or older were the group most consistently found at highest attack risk.

The results also point to what model developers can do to mitigate this risk. Stronger weight decay reduced the number of patients above the audit threshold with little additional GPU-board energy, and partially fine-tuning a foundation model reduced both attack scores and energy use while retaining similar diagnostic performance. Mixup and smaller training sets, by contrast, increased the fraction of patients above the attack threshold, showing that training choices can affect privacy in different directions. High diagnostic performance did not guarantee low attack risk either, as 45% and 90% of patients exceeded the risk threshold in the higher-performing ISIC-2020 and MRI-CORE, respectively. Once patients at risk are identified, removing their data is not the only possible response, as self-distillation reduced mean attack scores, with negligible impact on diagnostic performance. DataConfessor, the proposed data screening tool, may therefore be used to identify where that risk is concentrated before a production model is trained and apply a mitigation where one is warranted, but it should complement rather than replace formal privacy guarantees such as differential privacy[28,29].

Our study has limitations. All datasets were public research collections, so performance in independent hospital pipelines remains to be established. The 0.7 attack threshold is an operational, statistical threshold rather than a validated boundary for privacy harm. The normalized stability analysis may also overestimate reproducibility. The association between atypicality and audit score is not evidence of a causal mechanism, as both are computed from model confidence and depend on the labels, model capacity and optimization used. The screen identifies patients in a high-score tail and does not establish that unflagged patients are safe. On SCOPE-MRI, the smallest cohort, the highest-scoring 5% comprised 27 patients, so the screen's recall carries wide confidence intervals, and screening was evaluated at an earlier checkpoint because later scores approached saturation. Demographic comparisons were limited by the variables each dataset records: race and ethnicity were available only for CheXpert and SCOPE-MRI, and rarer groups were too small to exclude moderate differences. Finally, energy consumption was measured for the GPU-board only.

In conclusion, privacy screening before training is inexpensive, does not require the final production model,  and is available as open-source software. We propose that it become a routine step in preparing medical datasets, alongside the quality and bias checks that already precede model development.

# Methods

## Study design

For each dataset, we trained arms of shadow models, each model on a random half of the patients in the pool (Fig. 1a). An arm is a set of models that share a training strategy. Each reference arm has a seed replicate, which is the same recipe retrained with a different seed, which draws new patient halves and training seeds and measures chance variation. For example, the used NIH-CXR14 dataset held 19750 radiographs (to which we added 250 canaries, that are planted cases with shuffled labels, used to ensure that the attack itself worked) from 5,416 patients. In the reference arm, 128 models were trained on random halves of the patients and then scored every record, so each record's attack AUC compared about 64 models that had seen the patient (member) with 64 that had not (non-member). Each patient's risk was the maximum over their records, and 4,767 patients exceeded the exposure threshold of 0.7, which was fixed in advance at the lower bound of acceptable discrimination for clinical prediction models[20]. The weight-decay arm repeated this with weight decay $10^{-2}$, and the seed replicate with a different seed base.

## Datasets and pools

We used CheXpert[14] (frontal chest radiographs), NIH-CXR14[15] (frontal chest radiographs), ISIC-2020[16] (dermoscopy), PTB-XL[17] (12-lead, 10-s electrocardiograms), SCOPE-MRI[18] (shoulder MRI) and PneumoniaMNIST[19] (paediatric chest radiographs). Each record is one radiograph, dermoscopic image, electrocardiogram or MRI series. The pools contained 20,000, 20,000, 33,126, 21,799 and 3,815 records, from 6,860, 5,666, 2,056, 18,869 and 546 patients, respectively and PneumoniaMNIST contained 5,856 examples without patient identifiers, which were analysed individually (Figure 1). These numbers include, for each dataset, 250 planted records with randomized labels (canaries; 128 for SCOPE-MRI, as it was a smaller dataset), which served as a positive control (Validity checks for the attack). Canaries were excluded from every subsequent analysis. PneumoniaMNIST has no patient identifiers, so its examples were analysed individually. For SCOPE-MRI, 1,069 localizers and 150 duplicate reconstructions were excluded, and each remaining series was centre-cropped to a 78-mm field of view, resized to 224 pixels per slice and resampled to 24 slices. For CheXpert, race and ethnicity came from the Stanford AIMI demographic release, and insurance type from CheXpert Plus.

## Model training

Reference training strategies differed between datasets, following the most successful published architectures for each dataset, so every strategy comparison was made within a dataset, between sets of models (arms) with various training settings/hyperparameters. All reference recipes used weight decay $10^{-4}$. CheXpert, NIH-CXR14 and ISIC-2020 used an ImageNet-initialized ResNet-18 at 224 pixels for 10 epochs. PTB-XL employed a one-dimensional ResNet-18 (ResNet1D), PneumoniaMNIST a ResNet-18, both trained from random initialization for 50 epochs. The ISIC-2020 arms trained a ResNet-18 and an EfficientNet-B3 at 512 pixels on the full-resolution archive for 15 epochs. SCOPE-MRI models applied an ImageNet-initialized two-dimensional backbone (ResNet-18 or ViT-S/16) to each slice and max-pooled features across slices, following the approach of Sethi et al.[18], for 30 epochs. During architecture selection, this aggregation outperformed three-dimensional convolution and slice stacking. The SCOPE-MRI initialization comparison used random weights, ImageNet weights and ImageNet weights further pretrained on knee MR scans from MRNet[24]. Training modifications were first screened on PneumoniaMNIST (12 recipes; Fig. 2e), and weight decay $10^{-2}$ was then applied to all datasets. On PTB-XL, whose weight-decay arm failed the canary control, group normalization replacing batch normalization served as the contrast. To balance statistical robustness with computational feasibility, the number of shadow models was adapted to each experimental configuration: 128 models for most arms, 256 for PneumoniaMNIST and 32 for the 512-pixel ISIC-2020, SCOPE-MRI, PTB-XL group-normalization, mixed-precision and foundation-model arms. Each model's AUC was measured on the patients it was not trained on. The trained models were compared with a published model for the same dataset, when available, to set the utility floor, with a tolerance of 0.05 AUROC.

## Foundation models

ECGFounder[22] (Net1D, 30.7 million parameters, pretrained on about 10 million electrocardiograms) was fine-tuned from its released weights on PTB-XL with AdamW (learning rate $10^{-4}$, decoupled weight decay 0.01) for 20 epochs, with evaluation at epochs 1, 2, 5, 10 and 20. Full fine-tuning was the reference (30.7 million parameters fine tuned, 100%), compared against a seed replicate and a set of models with weight decay 1.0. Two sets of models trained only the last block of the last stage and the head (partial fine-tuning;

3.42 million parameters, 11%) or only the head (linear probe; 5,125 parameters, 0.017%). A control trained with Adam and L2 regularization failed the canary control. For each training (fine-tuning) variation, including seed replicates, 32 models were trained.

MRI-CORE[23] (ViT-B/16, 86.2 million parameters, pretrained on MRI) was fine-tuned from its released weights on SCOPE-MRI with AdamW (weight decay $10^{-4}$) for 20 epochs, with evaluation at epochs 1, 2, 3, 5, 8, 12, 16 and 20, and shared the dataset, canaries and membership of the ImageNet-initialized ResNet-18 reference ($N = 32$). Low-rank adaptation (LoRA[30]; rank 8, 0.44 million trainable parameters, 0.51%) had the best performance and served as the reference, with a seed replicate. The other arms were a linear probe (1,538 parameters), full fine-tuning (85.8 million parameters) and full fine-tuning with weight decay $10^{-2}$; the weight-decay contrast used full fine-tuning because strong weight decay could switch off the zero-initialized LoRA adapter. For each of these training (fine-tuning) variations, including seed replicates, 32 models were trained. The linear probe failed the canary control and was not analysed further.

### Membership inference and per-patient risk

A membership-inference attack infers from a model's outputs whether a record was in its training set, since models are usually more confident on their training records[2,6]. The attack and its evaluation follow established practice for membership-inference auditing[3,5]. For each model and record, confidence was the logit of the probability p assigned to the correct label, $\log(p/(1 - p))$, averaged over labels for multi-label tasks[3]. Each record was scored once. A record's attack AUC is the probability that a random model which trained on the patient (member), is more confident on it than a random model which did not train on him (non-member) (Mann–Whitney; 0.5, chance; 1.0, certain inference). Without leakage, member and non-member confidences would be interchangeable, and the AUC would follow an exact null distribution set only by the record's numbers of member and non-member models. The calibrated excess over null is the fraction of records with AUC above 0.7 minus the fraction expected under this null.

Patient risk was the maximum attack AUC over the patient's records[5], and patients with risk above 0.7 were considered exposed. At this threshold, a model trained on the patient is more confident about them than one not trained on them in 70% of pairs. The threshold affects exposed counts, the excess over null and the demographic analysis, but not normalized stability, Cohen's d or screen recall, which use ranks. Population-level attack AUC and the true-positive rate at a 0.1% false-positive rate used the online likelihood-ratio attack[3]. This attack fits normal distributions to each record's member and non-member confidences, leaving out the attacked model, and asks which better explains that model's confidence.

## Validity checks

No privacy measurement was reported for an arm unless it passed four checks: all records of a patient were either in or out of each model's training set; every record had at least four member and four non-member models. Canaries were considered identified/recovered, above a median AUC of at least 0.65 and higher AUCs than ordinary records (one-sided Mann–Whitney U test, $P \leq 10^{-6}$) and if a shuffled-label run reproduced the analytic null. Arms failing a check were excluded from every count.

## Matched model counts and checkpoints

The variability of a per-record AUC depends on the number of models, so arms were compared at matched $N$ by truncating each arm to its first N models ($N$ = 32–128). Counts and normalized stability used each recipe's final checkpoint (epoch 30 for SCOPE-MRI), at which the canary control passed. By then, SCOPE-MRI models had memorized most series, and many patients reached the maximum AUC, so on SCOPE-MRI the analyses that order patients within the tail (Cohen's d and screen recall) used epoch 5.

### Normalized stability

Normalized stability is S = ρ(A,B)/ρ(A,A′), where ρ is the Spearman correlation of per-patient risk, A and B are the reference and comparison arms and A′ is the seed replicate of A. S was capped at 1.0 and interpreted only where ρ(A,A′) > 0.5, and the threshold for invariance was S ≥ 0.8. The compared arms share the seed of their reference, whereas replicates do not. The ECGFounder partial-fine-tuning and linear-probe arms memorize little, so their patient rankings are noisy, and S understates their agreement with the reference. For them we report Spearman's ρ corrected for each arm's measurement noise, $\rho_c(A, B) = \frac{\rho(A,B)}{\sqrt{r_A r_B}}$, with r estimated from a seed replicate.

### Atypicality and candidate predictors

Atypicality is the negated mean confidence of models that did not train on a patient (non-member), in that patient record's correct label (the held-out component of the memorization score[8]), maximized over a patient's records. We compared it with four other predictors computed from the same models: the memorization gap (member minus non-member confidence), the lowest non-member confidence, the number of records per patient and held-out disagreement (the spread of non-member predictions). Each was scored by Cohen's $d$ between the 5% most exposed patients and the rest, with Benjamini–Hochberg correction[31] across predictors and datasets. Because patients with more records have a higher maximum risk by chance, effects were also computed within strata of record count and, to exclude shared estimation noise, from the seed-replicate arm.

### Screening tool

DataConfessor has been developed herein as a screening tool, which trains four proxy models for one epoch on halves of the dataset, so that each record is scored by two models that did not see it, and flags the 20% of patients with the highest atypicality. If the rankings of two independent proxy pairs have a Spearman correlation below 0.7, this triggers a fail-safe and increases the number of epochs for the proxy models. On SCOPE-MRI the correlation after one epoch lay at the threshold (0.68–0.73 across three runs), so five epochs were used. The proxies share only the data with the audited arms, so performance evaluation (recall) is prospective. One screen per dataset was scored against every arm trained on that dataset that passed the canary control. Recall is the fraction of an arm's top-5% most exposed tail, as defined by the full vulnerability assessment, that is contained in the flags (predictions of exposed patients by the tool). The package includes a command-line interface and a SKILL.md file that tells an AI assistant how to adapt the data loader to a new dataset, thus requiring no coding knowledge in order to be used.

### Tool's performance vs statistical baselines

The statistical baselines (predictors) were scored in the same way as the tool: the top 20% of patients were flagged and recall was computed against the same top-5% tail. Four families, 27–29 predictors per dataset, were compared: (A) metadata and raw-signal statistics, including record count, label rarity and the special-uniques disclosure score[25], (B) outlier scores on features from one forward pass of a frozen ImageNet ResNet-18 (spectral band powers on PTB-XL), (C) closed-form pixel statistics, such as texture, intensity histograms and sharpness, (D) unsupervised image models, such as principal-component reconstruction. Record count predicts the tail through order statistics alone (worse than atypicality) and served as the floor.

### Mitigation of vulnerability

To test whether patients at risk could be protected without removing their data, we retrained the reference recipes of ISIC-2020, PTB-XL and NIH-CXR14 with a mitigation strategy applied. Each defended arm used the same seeds and membership as the undefended (reference) arm, so the arms could be compared patient by patient. The first strategy was self-distillation. Two teacher models were trained with the same recipe on complementary halves of the patients, and the defended model was then trained on the probabilities predicted for each record by the teacher that had not seen its patient, instead of on the actual dataset label[27]. The second strategy used fixed softened labels (0.75 on the true class)[32,33] and the third used class-wise histogram matching. The primary endpoint was the mean change in risk of flagged patients relative to the undefended arm at the final checkpoint. A reduction was declared only if its bootstrap interval lay entirely below zero.

### Demographic analysis

The analysis was performed based on exposure (per-patient risk > 0.7) and membership of the top-5% tail (the 5% of patients with the highest risk). Both used arms that passed the canary control, which were assessed at the final checkpoint and at a matched number of models per dataset (128 for CheXpert and NIH-CXR14; 32 for ISIC-2020, PTB-XL and SCOPE-MRI). This gave 23 arms on four datasets (CheXpert 4, NIH-CXR14 4, ISIC-2020 7 and PTB-XL 8, three of them ECGFounder). On SCOPE-MRI, 95% of patients exceeded the exposure threshold, so only the top-5% tail was analysed, in 11 arms. For exposure (risk >

0.7), each group's share among exposed and non-exposed patients was compared by Fisher's exact test, with a crude odds ratio and, for non-age contrasts, an odds ratio adjusted for age from logistic regression. Holm correction was applied across the 23 reference-arm tests. For the top-5% tail, over-representation was the group's share of the tail divided by its share of the cohort, with 95% CIs from a patient bootstrap. In the reference arm, tail membership was also modelled by logistic regression: crude, then adjusted for disease status and number of records per patient, then additionally for atypicality percentile. Insurance contrasts were also adjusted for age band. On SCOPE-MRI, risk ratios were also computed within label strata. Race, ethnicity and insurance groups were compared with the majority group (White, non-Hispanic, private insurance), age groups with the rest of the cohort (≥80 vs <80 y, 60–79 vs <60 y, <40 vs ≥40 y), men with women and head and neck lesions with other sites. Groups with fewer than 30 patients were omitted.

### Energy measurement

GPU energy was read from the board's built-in energy counter and reported with the CEDARS AI Research Label[34,35], using the 2026 carbon intensity of the Greek grid[36,37].

### Statistics

Uncertainty was estimated by resampling patients 2,000 times (patient bootstrap). For differences between arms trained on the same pool, both arms used the same resample. Differences in held-out AUROC were estimated across an arm's trained models with Welch's t-test. Multiple comparisons were corrected with the Benjamini–Hochberg procedure. Tests used for a single figure are described in its legend.

### Computational resources

Models were trained on a 16-GB Quadro RTX 5000 (PyTorch 2.5.1, CUDA 12.1) and, for SCOPE-MRI, a 32-GB RTX 5090 (PyTorch 2.14, CUDA 13.0). The study comprised 7,789

trained models (7,152 on the RTX 5000 and 637 on the RTX 5090), 1,420 GPU-hours (1,123 and 297) and 233–236 kWh.

### Use of large language models

During the preparation of this manuscript, local open-source models (llama3.1, qwen3.5) were used for language polishing and text formatting. The authors reviewed and edited the content as needed and take full responsibility for the final publication.

## Data availability

All six datasets are publicly available for research: CheXpert, NIH-CXR14, ISIC-2020, PTB-XL and PneumoniaMNIST from their providers, and SCOPE-MRI from the Medical Imaging and Data Resource Center under its data use agreement. CheXpert demographic and insurance variables are available from Stanford AIMI under its research-use agreement.

## Code availability

DataConfessor (screening tool) and the analysis code are available at https://github.com/konstvr/DataConfessor .

## Funding

Funded by the European Union under the Horizon grant agreement No. 101287679. Views and opinions expressed are however those of the author(s) and do not necessarily reflect those of the European Union nor European Research Executive Agency (REA). Neither the European Union nor REA can be held responsible for them.